\documentclass[conference]{IEEEtran}
\IEEEoverridecommandlockouts
\usepackage{cite}
\usepackage{amsmath,amssymb,amsfonts}
\usepackage{algorithmic}
\usepackage{graphicx}
\usepackage{subcaption}
\usepackage{textcomp}
\usepackage{xcolor}

\def\BibTeX{{\rm B\kern-.05em{\sc i\kern-.025em b}\kern-.08em
    T\kern-.1667em\lower.7ex\hbox{E}\kern-.125emX}}

\begin{document}

\title{Decentralized Flood Forecasting Using Deep Neural Networks
}

\author{\IEEEauthorblockN{1\textsuperscript{st} Muhammed Ali Sit}
\IEEEauthorblockA{\textit{Department of Computer Science} \\
\textit{University of Iowa}\\
Iowa City, United States \\
muhammedali-sit@uiowa.edu}
\and
\IEEEauthorblockN{2\textsuperscript{nd} Ibrahim Demir}
\IEEEauthorblockA{\textit{Department of Civil and Environmental Engineering} \\
\textit{University of Iowa}\\
Iowa City, United States \\
ibrahim-demir@uiowa.edu}
}

\maketitle

\textit{This article is a pre-print submitted to ArXiv.}\hfill \break

\begin{abstract}
Predicting flood for any location at times of extreme storms is a longstanding problem that has utmost importance in emergency management. Conventional methods that aim to predict water levels in streams use advanced hydrological models still lack of giving accurate forecasts everywhere. This study aims to explore artificial deep neural networks' performance on flood prediction. While providing models that can be used in forecasting stream stage, this paper presents a dataset that focuses on the connectivity of data points on river networks. It also shows that neural networks can be very helpful in time-series forecasting as in flood events, and support improving existing models through data assimilation.
\end{abstract}


\section{Introduction}
2017 was the costliest year on record for natural disasters in the United States. Hurricanes Harvey \cite{van2017attribution}, Irma \cite{sit2019identifying} and Maria, and other natural disasters caused more than \$306 billion. Flood forecasting or stage height prediction is a problem that is being explored by hydrological models that requires years of experience and knowledge. Even though, years of work is invested in those advanced models they still lack in forecasting floods accurately at arbitrary locations and durations. 

In this study, we present a benchmark dataset for flood forecast to generate and test deep neural network performance for accurate prediction of flood prediction. Since the flood forecasting is a task involves time-series data, both dataset and networks are created taking this into account.

This paper first discusses related works in the next section. Section II provides preliminary information regarding the task and raw data. The dataset creation process is scrutinized in detail in the same section, after that the proposed neural network structures are introduced. In the results and discussion section, outcomes of the performance metrics are shared for proposed networks and the results are provided before the conclusion section.

\subsection{Related Work}

Neural networks were widely used in the past for hydrological modeling purposes. \cite{dawson1998artificial,thirumalaiah2000hydrological,dawson2001hydrological,nayak2005short,chang2007multi} Most of the studies extend the feed-forward networks and propose a neural network model that does not involve a high number of dataset entries because of the technical limitations. Also, these studies involve mostly fuzzy logic instead of artificial intelligence focused aspects of deep neural networks. There are studies that utilizes artificial neural networks for flood forecasting purposes as well \cite{campolo1999river,huang2004forecasting,lin2008systematic,liu2017flood}. These studies present backpropagation networks centric approaches in flood forecasting. Also, in \cite{lin2004non,lin2011rbf,chaowanawatee2012implementation}, radial basis function neural networks were employed in streamflow prediction. Machine learning algorithms other than neural networks are widely explored in the literature. \cite{han2007flood, yu2006support} explore Support Vector Machines in flood forecasting. Besides limited studies in flood forecasting and gage height prediction, deep neural networks are utilized with latest algorithmic advances for similar tasks in hydrology such as reservoir inflow forecasting \cite{bai2016daily}, precipitation estimation \cite{tao2016deep}, runoff analysis \cite{izumirunoff}.  Even though forecasting is mostly time depended, forecasting tasks mentioned here vastly comprises network architectures that do not take advantage of sequential nature of time series data. Flood forecasting also involves time-series data and overcoming this task should be done by creating the dataset centered upon its time-depended features.

Studies that use deep neural networks for time-series data show great results and provides an extensive vision in increasing the usage of neural networks architectures in time-series data. Researchers show that neural networks can be used for traffic speed prediction \cite{wang2016traffic, zhang2017deep, jia2016traffic}, taxi demand prediction \cite{yao2018deep}, financial time-series prediction \cite{ding2015deep,bao2017deep}. Traffic speed prediction is a very similar task with flood prediction since both of them relies on changes in connected points in the network.

In this paper, we propose a flood prediction benchmark dataset for future applications in machine learning as well as a scalable approach to forecasting river stage for individual survey points on rivers. The approach takes into account the historical stage data of survey points on selected upstream locations, as well as precipitation data. This approach is both decentralized and doesn't need any historical data from unrelated survey points and can be used in real-time. Recurrent Neural Networks (RNNs), in particular, Gated recurrent unit (GRU) Networks are utilized throughout this study. We also show that this approach presents satisfiable results when it's applied for the state of Iowa as a proof of concept. The data are gathered from the Iowa Flood Center (IFC) and United States Geological Survey (USGS) sensors on the rivers within the state of Iowa. Findings of this project and the deep neural network model will benefit operational cyber platforms \cite{demir2017floodss}, intelligent knowledge systems \cite{sermet2018intelligent}, and watershed information and visualization tools  \cite{demirdatafield,demirscivis,demirgwis} with enhanced river stage forecasting.

\section{Methodology}
In this section methods that this study relies on are explained. Preliminary information regarding the problem definition and details about the case study dataset is detailed below. After the dataset information, deep neural network models is presented.

\subsection{Preliminary Work}

Flood forecasting or prediction of river stage for a particular point on the river network depends on the neighboring streams. The problem of flood forecasting should be solved by anticipating the height of the water on streams. This problem and potential physical approaches expect a deep understanding of the hydrology and domain.

Since the aim of this study is to predict the water level for selected points on rivers, river network structure and connectivity should be understood by the framework. If the intersection points on river network are considered as nodes and rivers are considered as edges, river networks make directed acyclic graphs (DAGs). Water only flows from upstream to downstream and it never makes circular moves. In this study, instead of using intersections, gage height sensors are used as nodes. United States Geological Survey (USGS), has gage height sensors all around the US. Sensors provide gage height measurements with a temporal resolution of 15 minutes. By using these sensors, a sensor network can be formed. The river network is a DAG, the existence of more than one sensor on a river is possible, and some nodes can have only one child node. Each sensor has upstream sensors and downstream sensors. Among two connected sensors, or nodes, A and B, if the water in their watershed reaches the sensor A before it reaches to the sensor B, then it’s clear that A is on B’s upstream and B is on A’s downstream. The water level of a sensor vastly relies on the water level of its upstream sensors. Therefore, it’s important to incorporate upstream water level information into the forecasting effort.

As a data-driven approach to testing the capabilities of deep learning for mapping rainfall and runoff, this study doesn't involve other mechanisms and processes that affect gage heights and floods.

\subsection{Dataset}

Data used in this study comprises 3 different sets. The first dataset consists of gage height from USGS and UFC sensors. The second dataset is the NOAA’s Stage IV radar rainfall product and the last dataset is the metadata and sensor information from IFC. USGS has 201 gage height sensors within the state of Iowa, each providing gage height measurements with 15 minutes temporal resolution. Data obtained from USGS data sources for these sensors contain the sensor id, date and time of the measurement with timezone, and the measurement value. We needed to preprocess these data in order to form a data structure for our models. The preprocessed dataset was created as a hash table that has sensor ids as keys and has the value of another hashtable of datetime, measurement dates. The internal hashtables that have datetime as their keys were converted to datetimes in UTC which was previously CST or CDT depending on measurement’s the time of the year to avoid any timezone related problems.

Another preprocessing step was done to extract usable sensors. The IFC data regarding sensors contains network information for the sensors. We extracted the upstream sensor list for each USGS sensor. The list for each sensor was sorted by their proximity to the intended sensor. The number of upstream sensors that will take place on the input data was selected to be 4 and when the sensors that have a number of upstream sensors lower than that value was discarded. The final usable USGS sensor list contained 45 sensors. Values other than 4 was tested but no significant improvement was observed.

\begin{figure}
\centerline{\includegraphics[width=0.5\textwidth]{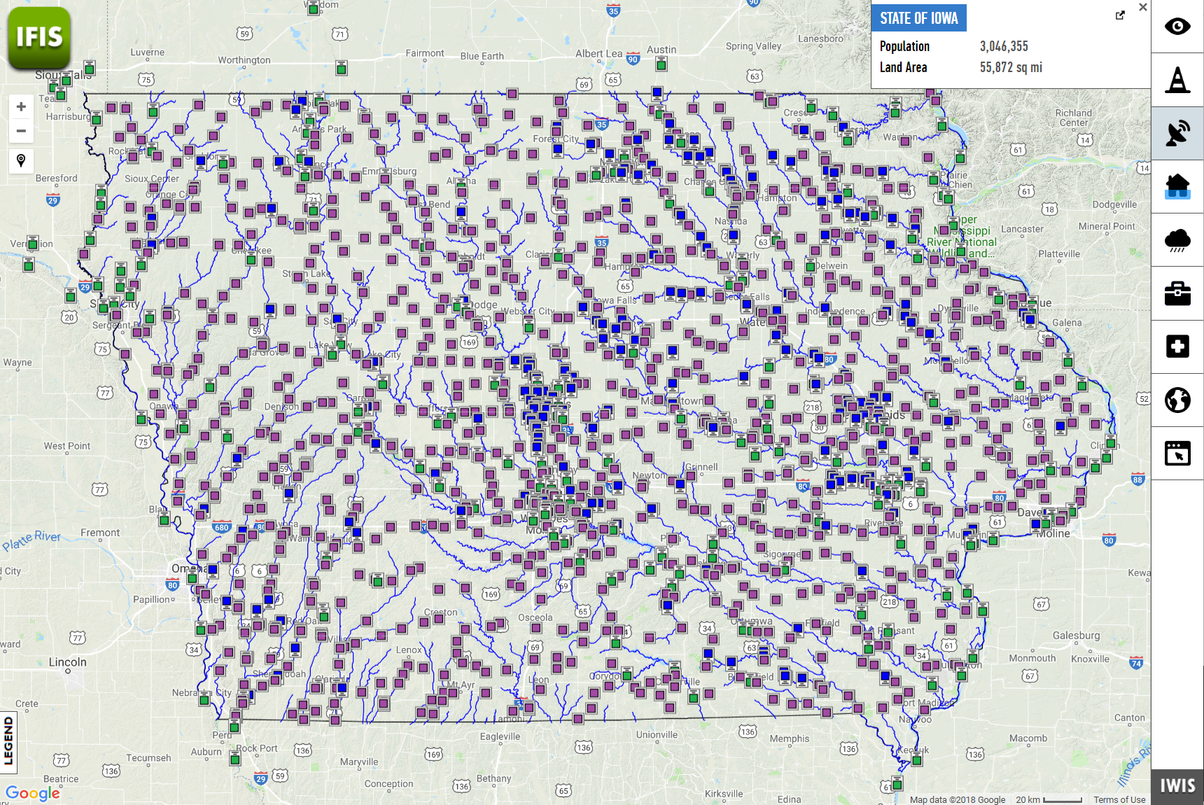}}
\caption{USGS and IFC stream sensors in Iowa}
\label{fig}
\end{figure}

For a dataset entry with a sensor \textit{S} and time \textit{t}, the input part of the dataset vastly comprises gage height data from upstream sensors of the sensor \textit{S} for measurements before time \textit{t}, as well as the previous measurements from  the sensor \textit{S} itself. Also, rainfall that falls within the watershed of sensor \textit{S} takes place in the input part of the data. Output part of the data is future measurements of sensor \textit{S} after the time \textit{t}. In this study, the approach is designed to predict 24 hours of future measurements thus the output consists of 24 values, in which, one value represents each forecasted hour.

We formed two similar datasets using the preprocessed data. They differ in size for the historical information they have regarding sensors and precipitation. The smaller dataset only has 4 hours of stage measurements for each upstream sensor and 3 hours of information for the intended sensor as well as precipitation data array with the length of 20. The larger dataset has 24 hours of data for each upstream sensor, and 24 hours of data for the intended sensor and precipitation data array with the length of 40. The output parts for both datasets were same.

The datasets were formed by creating a dataset entry vector for each 15 minutes datetime instance between January 2009 and June 2018. While each vector for smaller dataset comprised total of 39 input values and 24 output values, vectors of the larger dataset had a total of 160 input values and 24 output values. Since each input has two parts (height and precipitation data), we formed them separately and then concatenated them together. For clarification purposes, we will explain the dataset using the same approach.

  \begin{equation}
  \begin{aligned}
H = [H^{U^1, S}_{t-td-3}, H^{U^1, S}_{t-td-2}, H^{U^1, S}_{t-td-1}, H^{U^1, S}_{t-td},\\
H^{U^2, S}_{t-td-3}, H^{U^2, S}_{t-td-2}, H^{U^2, S}_{t-td-1}, H^{U^2, S}_{t-td}, \\
H^{U^3, S}_{t-td-3}, H^{U^3, S}_{t-td-2}, H^{U^3, S}_{t-td-1}, H^{U^3, S}_{t-td}, \\
H^{U^4, S}_{t-td-3}, H^{U^4, S}_{t-td-2}, H^{U^4, S}_{t-td-1}, H^{U^4, S}_{t-td}, \\
H^{S}_{t-3}, H^{S}_{t-2}, H^{S}_{t-1}]
\label{heightvector}
 \end{aligned}
\end{equation}

Height vector, \textit{H} (Equation \ref{heightvector}, for smaller dataset in which $H^{U^2, S}_{t-td-2}$ represents height data for second sensor in upstream of \textit{Sensor S} at time \textit{t-td-2}.) part of any dataset entry was created using USGS data. When gathering data for an output of 24 hours starting at time \textit{t}, for each upstream sensor, data were taken for $t-td$ in which \textit{td} refers the travel time distance of water between the upstream sensor and the sensor that is intended to anticipate stage height change. Time distance information was extracted from the sensor data by IFC. Each sensor point has time distance to the outlet point in the corresponding watershed. By taking the time distance difference between an upstream sensor and the intended sensor, the time distance between these two sensors can be calculated. Using this information, incorporated upstream data included the data for the measure water level that most likely will affect the intended sensor's water level.

\begin{figure}
\centerline{\includegraphics[width=0.5\textwidth]{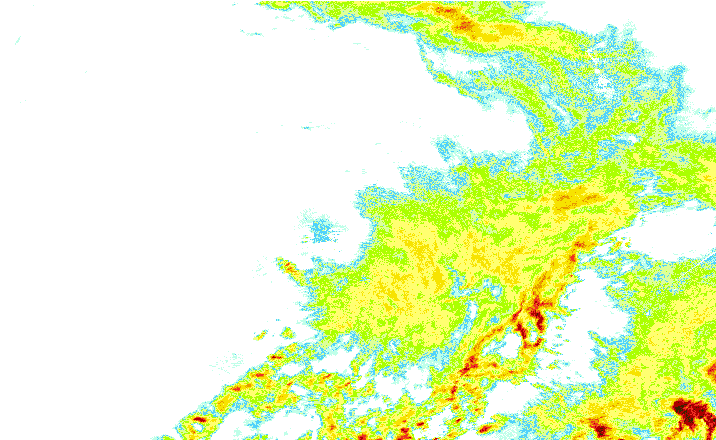}}
\caption{Sample precipitation data of the selected rainfall domain for the State of Iowa.}
\label{rainfall}
\end{figure}

The second vector that comprises a dataset entry is precipitation vector. Rainfall data from Stage IV product (Fig. \ref{rainfall}.) provided in rasters. The data in raster files contains precipitation values for the entire US divided into parcels. After converting raster files into easily accessible arrays, the next step is to determine the approach that will be used when acquiring the data from the rasters for each sensor and datetime pairs. The first approach was to use precipitation data only for parcels that include exact locations of upstream sensors and the intended sensors without considering their watersheds but this approach doesn't take the rainfall that falls to the area between sensors into account. Another approach was to use data for the entire watershed of the intended sensor. Even though this approach has the potential to represent the domain better, since all of the precipitation in upstream sensor watersheds drains and are represented in the gage measurements, this approach brings unnecessary data amplitude. 

\begin{figure}
\centerline{\includegraphics[width=0.5\textwidth]{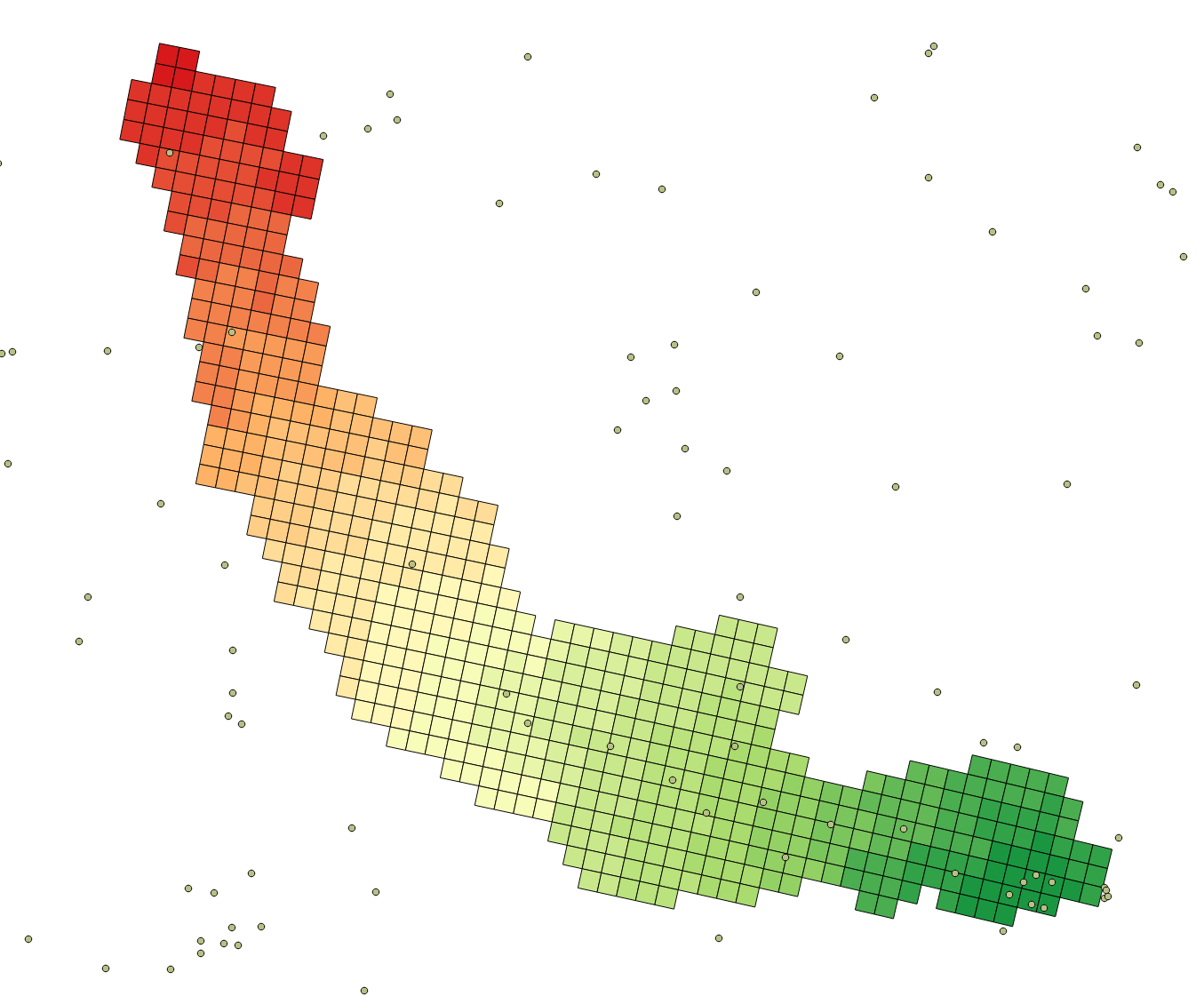}}
\caption{Watershed of a sensor with parcels colored depending on their time distance to the intended sensor (green is closer)}
\label{watershed}
\end{figure}

In order to both keep important information in the dataset and skip the already obtained information, we used an approach in which the sections of the watershed area is eliminated based on upstream sensors. Watershed area that will be used to gather precipitation data representing rainfall domain that will eventually reach intended sensor is calculated, and the watershed area of upstream sensors was excluded from the watershed of the intended sensor. Remaining parcels were divided into sections depending on the parcels' water time distance to the actual sensor (Fig. \ref{watershed}.).

After gathering the watershed information that will be used to acquire rainfall data, the rasters were read and depending on the time distance, precipitation measurements were obtained. For instance if the output vector contains measurements for the sensor \textit{S} at time \textit{t}, the precipitation data would be obtained for $ t-td $ for each parcel. Then, the average of the measurements that parcels have was taken for each time distance value. Eventually, all of the precipitation vectors were formed but they almost always had a different length. To provide the same length vectors, the precipitation vectors were padded into vectors with a length of 20 or 40 depending on the aforementioned dataset sizes. If the vector is larger than the size predetermined vector size, the remaining part was cropped out and if the vector was smaller than the size, empty places were filled with zeros. Equation \ref{precvector} demonstrates final precipitation vector \textit{P} for smaller dataset in which $P_{t-8}^{S}$ represents mean of precipitation data for parcels that are 8 hours away from the sensor \textit{S}.

  \begin{equation}
  \begin{aligned}
P = [\;\;\!P_{t-1}^{S},\;\;\! P_{t-2}^{S},\;\;\! P_{t-3}^{S},\;\;\! P_{t-4}^{S},\;\;\! P_{t-5}^{S},\; \\
P_{t-6}^{S},\;\;\! P_{t-7}^{S},\;\;\! P_{t-8}^{S},\;\;\! P_{t-9}^{S},\;\;\! P_{t-10}^{S}, \\
P_{t-11}^{S}, P_{t-12}^{S}, P_{t-13}^{S}, P_{t-14}^{S}, P_{t-15}^{S}, \\
P_{t-16}^{S}, P_{t-17}^{S}, P_{t-18}^{S}, P_{t-19}^{S}, P_{t-20}^{S}\;]\\
\label{precvector}
 \end{aligned}
\end{equation}

The last step was to form output vector. Output vector for a dataset entry of a sensor \textit{S} that will contain data for time \textit{t} will have height measurements from \textit{t} to \textit{t+23}. Equation \ref{outputvector} shows the form of output vector \textit{O} while $H^{S}_{t+17}$ represents height measured on Sensor \textit{S} at time \textit{t+17} in feet.

  \begin{equation}
  \begin{aligned}
O = [H^{S}_{t},\;\;\;\;\;\!H^{S}_{t+1},\;\;\!H^{S}_{t+2},\;\;\!H^{S}_{t+3},\;\\
H^{S}_{t+4},\;\;\!H^{S}_{t+5},\;\;\!H^{S}_{t+6},\;\;\!H^{S}_{t+7},\;\\
H^{S}_{t+8},\;\;\!H^{S}_{t+9},\;\;\!H^{S}_{t+10},H^{S}_{t+11},\\
H^{S}_{t+12},H^{S}_{t+13},H^{S}_{t+14},H^{S}_{t+15},\\
H^{S}_{t+16},H^{S}_{t+17},H^{S}_{t+18},H^{S}_{t+19},\\
H^{S}_{t+20},H^{S}_{t+21},H^{S}_{t+22},H^{S}_{t+23}\;]\\
\label{outputvector}
 \end{aligned}
\end{equation}

After both precipitation and height vectors were formed, the input vector was created by concatenating them. Both smaller and larger datasets were formed by this pipeline only differing in the sizes when applicable. After running through this process for each sensor and datetime instance within the mentioned date range, 298,496 dataset entries were formed for the larger dataset and 354,816 dataset entries were formed for the smaller dataset. Recall that smaller and larger names are used depending on the size of the individual input vectors, not the actual dataset size.

\subsection{Utilizing Neural Networks}
Predicting stream heights vastly depends on previous states of streams and this makes the flood forecasting problem more of a time-series forecasting task. Considering this nature of the problem, using more sequentially capable network architectures like RNNs make good implementation choices. Due to vanilla RNN networks' vanishing gradient problem, Long short-term memory (LSTM) networks and GRU networks are major two network architecture options in the literature for such cases. Since GRU networks have less computationally costly operations than LSTM networks, in other words, because of GRU has fewer gate computations but still matches the LSTM's performance, we chose to implement a GRU based network. 

\begin{equation}
  \begin{aligned}
f(x) = max(x, 0)
\label{relu}
 \end{aligned}
\end{equation}

This study proposes two networks for comparison purposes, the first one is fully-connected network and the second one is the GRU based neural network. The fully-connected network structure can be found in (Table \ref{fctable}). All layer outputs are activated with Rectified Linear Unit (ReLU) (\ref{relu}) function.

\begin{figure}
\centerline{\includegraphics[width=0.5\textwidth]{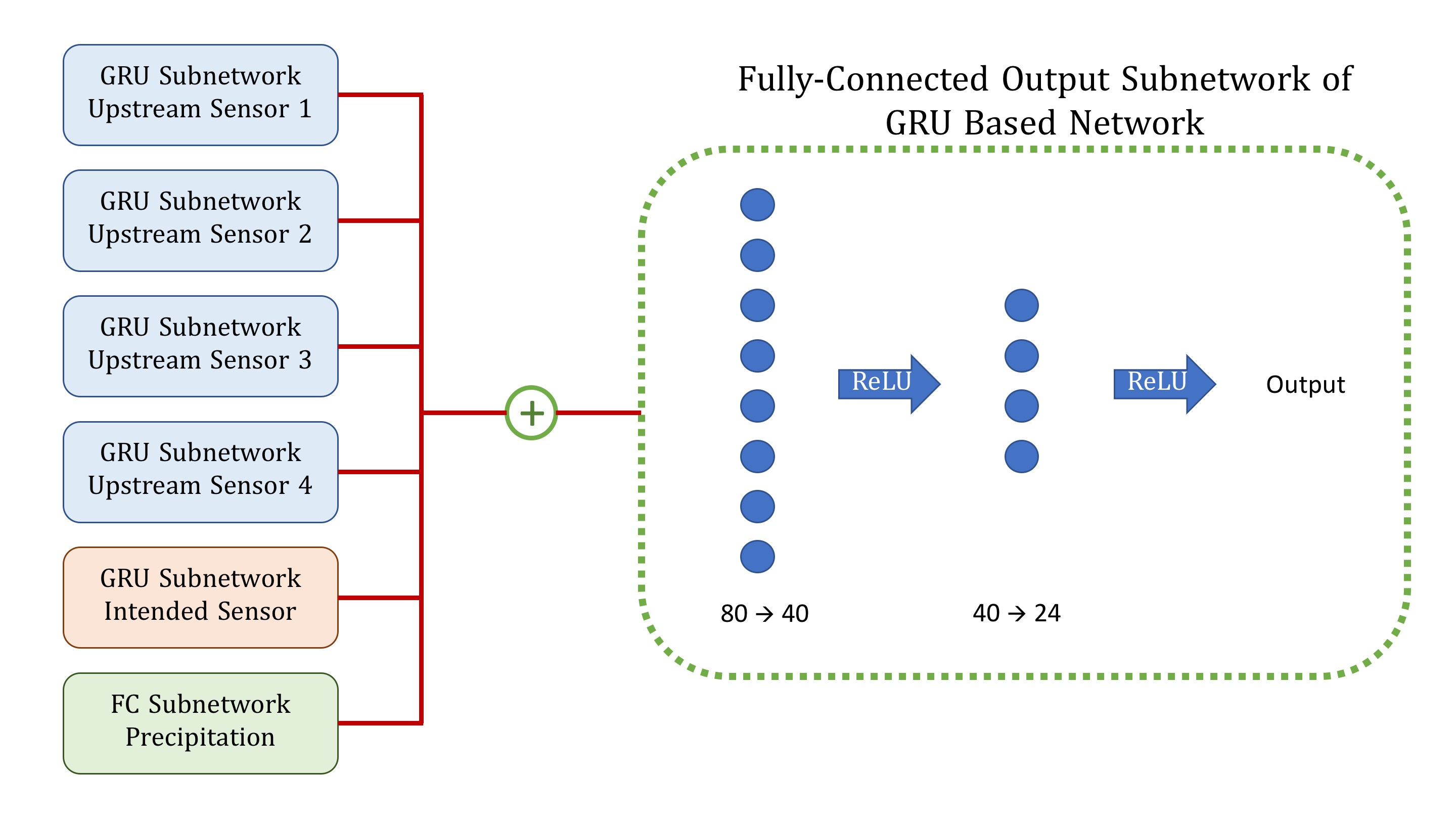}}
\caption{GRU-based network architecture.}
\label{fig_gru}
\end{figure}

GRU based network consists of five GRU subnetworks (\ref{subgru}), one for each upstream and one for previous measurements of the intended sensor, and a fully-connected subnetwork (Table \ref{subfc}) for the precipitation data. Outputs of all these subnetworks are then fed into a fully-connected output network with ReLU activations until the output (Figure \ref{fig_gru}) is computed.

Cho \textit{et al.}\cite{cho2014learning} in 2014 proposed GRU networks. GRU cells are similar to LSTM cells in terms of utilization and their capabilities in handling vanishing gradient problem. GRU comprises concepts that are easier to implement while providing very similar performance with the LSTM. A GRU cell has two gates, a reset gate in which previously learned features and new inputs are combined, and the update gate which determines how much of the memory will be remembered. A GRU cell's formulation can be expressed as,

\begin{equation}
r_t = \sigma(U^rX_t + W^rs_{t-1} + b^r)
\end{equation}
\begin{equation}
z_t = \sigma(U^zX_t + W^zs_{t-1} + b^z)
\end{equation}
\begin{equation}
h = g(U^hX_t + r_t \cdot (W^hs_{t-1}) + b^h)
\end{equation}
\begin{equation}
s_t = z_t \cdot s_{t-1} + (1-z_t) \cdot h
\end{equation}

where $r_t$, $z_t$, $h$ and $s_t$ represent reset gate, update gate, hidden state candidate and hidden state respectively. Also $\sigma$ is sigmoid function and $g$ is tanh function. In GRU cells, while reset gate affects the hidden state by taking place in hidden state candidate's formula, update gate significantly changes hidden state.

\begin{table}[htbp]
\caption{Fully-connected network architecture}
\begin{center}
\begin{tabular}{|c|c|c|}
\hline
\textbf{Layer} & \textbf{Input} & \textbf{Output} \\
\hline
Input Layer & 39 & 350\\
\hline
Fully-Connected & 350 & 500 \\
\hline
Fully-Connected & 500 & 350 \\
\hline
Output Layer & 350 & 24 \\
\hline

\end{tabular}
\label{fctable}
\end{center}
\end{table}

\begin{table}[htbp]
\caption{GRU sub-network architecture for GRU-based network}
\begin{center}
\begin{tabular}{|c|c|c|}
\hline
\textbf{Layer} & \textbf{Input} & \textbf{Output} \\
\hline
Input Layer & 24 & 400\\
\hline
GRU & 400 & 400 \\
\hline
Tanh & 400 & 400 \\
\hline
2D MaxPooling & 400 & 400 \\
\hline
Output Layer & 400 & 10 \\
\hline

\end{tabular}
\label{subgru}
\end{center}
\end{table}

\begin{table}[htbp]
\caption{Fully-connected sub-network architecture for GRU based network}
\begin{center}
\begin{tabular}{|c|c|c|}
\hline
\textbf{Layer} & \textbf{Input} & \textbf{Output} \\
\hline
Input Layer & 40 & 100\\
\hline
Fully-Connected & 100 & 200 \\
\hline
Fully-Connected & 200 & 100 \\
\hline
Output Layer & 100 & 30 \\
\hline

\end{tabular}
\label{subfc}
\end{center}
\end{table}

\section{Results \& Discussion}
\subsection{Results}
Proposed neural network architectures are implemented using PyTorch \cite{paszke2017automatic} numeric computing library (v0.40) and master version of experiment time (0.5.0a0+290d20b) on Python programming language (v3.6). The source code was written to train networks using the Adam Optimizer \cite{kingma2014adam} as the optimization method and mean squared error (MSE) as the loss function. Proposed datasets were split into training and testing sets with an approximate rate of 80\%. Implemented networks were trained on training sets using NVIDIA Tesla K80 GPUs. 

\begin{figure}
\begin{subfigure}{.5\textwidth}
  \centering
  \includegraphics[width=.9\linewidth,height=.21\textheight]{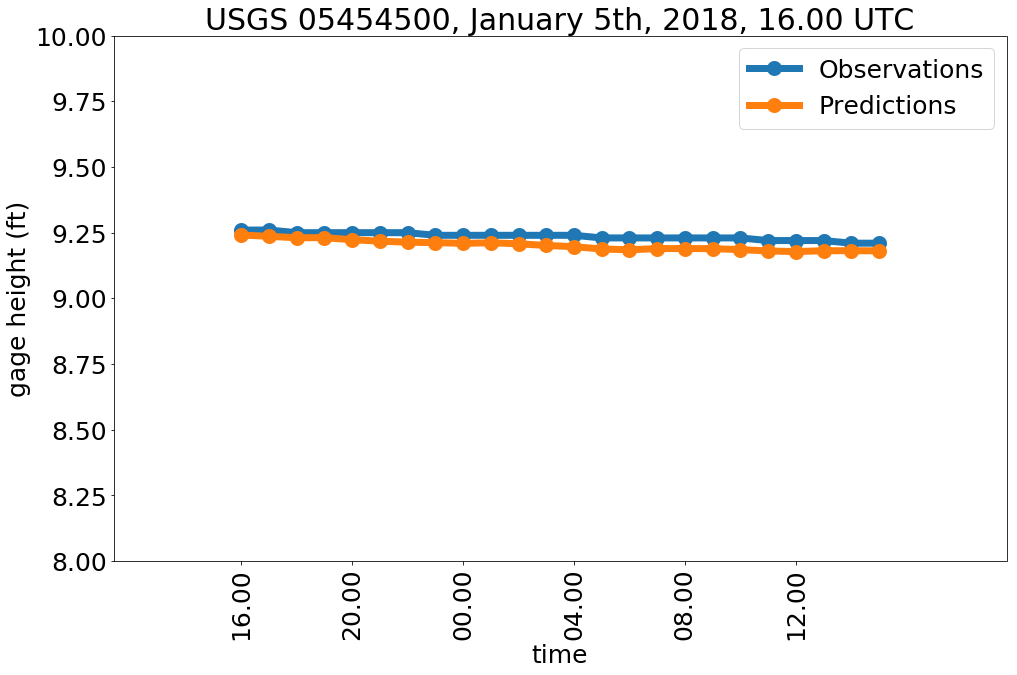}
  \caption{}
  \label{fig:sfig1}
\end{subfigure}
\begin{subfigure}{.5\textwidth}
  \centering
  \includegraphics[width=.9\linewidth,height=.21\textheight]{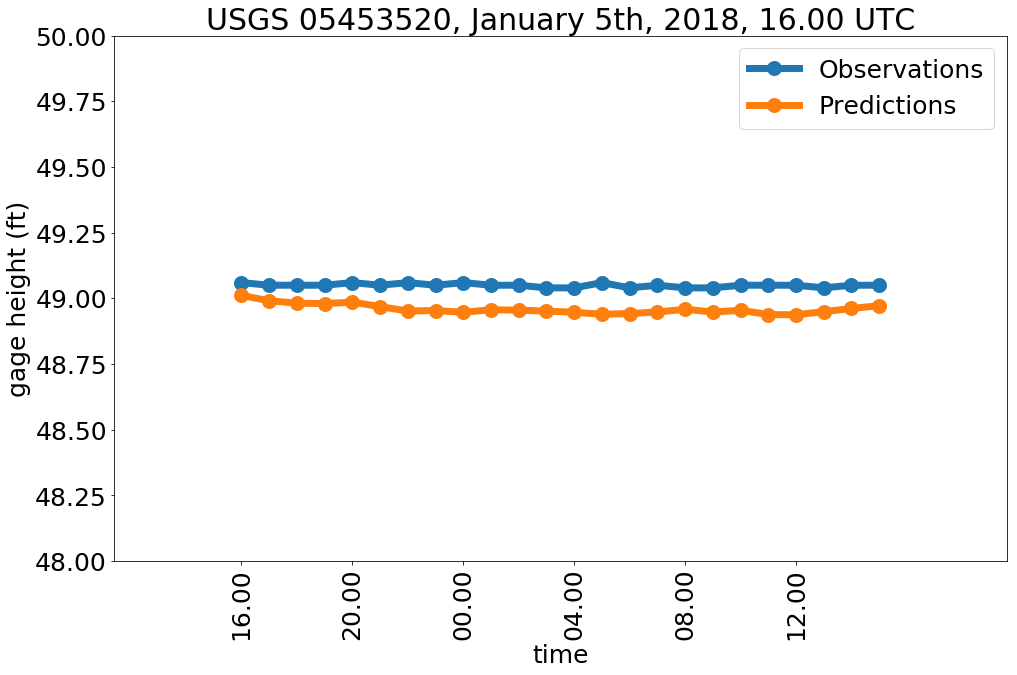}
  \caption{}
  \label{fig:sfig2}
\end{subfigure}
\begin{subfigure}{.5\textwidth}
  \centering
  \includegraphics[width=.9\linewidth,height=.21\textheight]{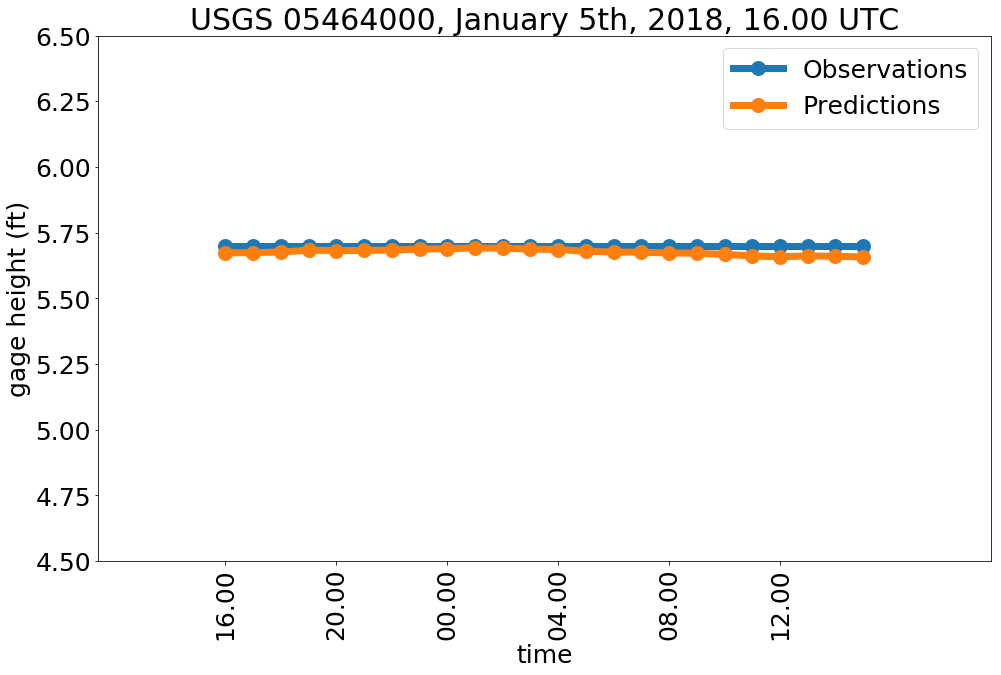}
  \caption{}
  \label{fig:sfig2}
\end{subfigure}
\begin{subfigure}{.5\textwidth}
  \centering
  \includegraphics[width=.9\linewidth,height=.21\textheight]{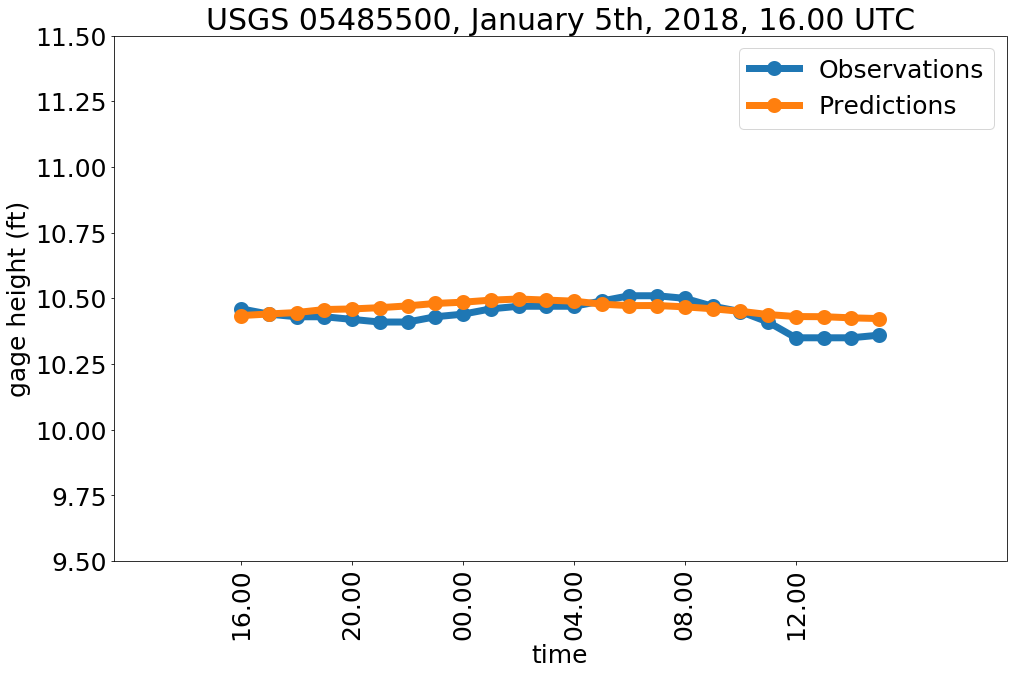}
  \caption{}
  \label{fig:sfig2}
\end{subfigure}
\caption{Gage height measurement values and predictions that GRU based model generated for 4 sensors}
\label{fig:results}
\end{figure}

It should be noted that while the larger dataset was run with both GRU-based architecture and fully-connected architecture, the smaller dataset was only run with fully-connected architecture to understand the effect of using less time-dependent data in such learning task.

\begin{table}[htbp]
\caption{Mean squared error scores for test sets (ft/h)}
\begin{center}
\begin{tabular}{|c|c|c|}
\hline
\textbf{} & \textbf{Larger Dataset} & \textbf{Smaller Dataset} \\
\hline
Fully-connected & 0.74 & 0.74\\
\hline
GRU Based & 0.60 & N/A \\
\hline
\end{tabular}
\label{mse}
\end{center}
\end{table}

Score table that shows MSE on testing datasets for proposed models which are trained on training datasets can be found in Table \ref{mse}. Scores clearly show that providing more data did not significantly improve the accuracy of the fully-connected model. However, it can be seen that model choice has important effects on the model's overall testing performance. We can easily say that RNN based models make better architecture choices for prediction tasks such as flood forecasting.

Actual measurements and values that GRU based network predicted for 4 USGS sensors are given in Figure \ref{fig:results}. Shared results suggest that when stage height does not show dramatic changes, the model is successful to anticipate next measurements but when there are apparent fluctuations, the model is not able to perform as successful, but still it reports somewhat similar predictions. It should be noted that even though the forecasts seem to not frequently show an exact match with actual measurements, they do not possess huge numeric differences.

Considering the reported MSE and the similarity between measurements and forecasts that GRU based model made, it can be said that the overall performance of the neural networks with the proposed decentralized approach is acceptable.

\section{Conclusion}
While this paper demonstrates a benchmark dataset and methodology for flood forecasting that employs deep neural networks, it also presents promising results using a data-driven approach. The approach in this study could be improved in the future by incorporating other datasets such as soil moisture data from point source measurements and satellite data such as Soil Moisture Active Passive (SMAP) as well as evaporation measurements which can help in demonstrating the water budged better.

Models proposed in this study can be used to present more enhanced forecasting results on operational information systems along with forecasts of advanced hydrological models. Presented results show that artificial neural networks based decentralized flood forecasting approach for the state of Iowa anticipates the stage height very close to the actual height measurements.

\section*{Acknowledgment}

The work reported here has been possible with the support and work of many members of the Iowa Flood Center at the IIHR Hydroscience and Engineering, University of Iowa.

\bibliography{mwe} 
\bibliographystyle{ieeetr}

\end{document}